\documentclass[10pt,twocolumn,letterpaper]{article}

\usepackage[pagenumbers]{iccv} 

\definecolor{iccvblue}{rgb}{0.21,0.49,0.74}
\usepackage[pagebackref,breaklinks,colorlinks,allcolors=iccvblue]{hyperref}
\usepackage{blindtext}
\usepackage{microtype}
\usepackage{multirow}
\usepackage{amsthm}
\usepackage{graphicx}
\usepackage{subcaption}
\usepackage{booktabs, arydshln}
\usepackage{nicematrix,tikz}
\usepackage{tabularx}
\usepackage[utf8]{inputenc} 
\usepackage[T1]{fontenc}    
\usepackage{hyperref}       
\usepackage{url}            
\usepackage{amsfonts}       
\usepackage{nicefrac}       
\usepackage{microtype}      
\usepackage{xcolor}         
\usepackage{amsmath}
\usepackage{algorithm}
\usepackage{algorithmic}

\usepackage{adjustbox}
\usepackage{array}

\newcolumntype{R}[2]{%
    >{\adjustbox{angle=#1,lap=\width-(#2)}\bgroup}%
    l%
    <{\egroup}%
}
\newcommand*\rot{\multicolumn{1}{R{90}{1em}}}

\def\confName{ICCV}
\def\confYear{2025}

\title{Optimizing Specific and Shared Parameters for Efficient Parameter Tuning}

\author{Van-Anh Nguyen\\
Monash University\\
{\tt\small van-anh.nguyen@monash.edu}
\and
Thanh-Toan Do \\
Monash University\\
{\tt\small toan.do@monash.ede}
\and
Mehrtash Harandi \\
Monash University\\
{\tt\small Mehrtash.Harandi@monash.edu}
\and
Dinh Phung \\
Monash University\\
{\tt\small dinh.phung@monash.edu}
\and
Trung Le \\
Monash University\\
{\tt\small trunglm@monash.edu}
}

\begin{document}
\maketitle
\begin{abstract}
Foundation models, with a vast number of parameters and pretraining on massive datasets, achieve state-of-the-art performance across various applications. However, efficiently adapting them to downstream tasks with minimal computational overhead remains a challenge. Parameter-Efficient Transfer Learning (PETL) addresses this by fine-tuning only a small subset of parameters while preserving pre-trained knowledge.
In this paper, we propose SaS, a novel PETL method that effectively mitigates distributional shifts during fine-tuning. SaS integrates (1) a shared module that captures common statistical characteristics across layers using low-rank projections and (2) a layer-specific module that employs hypernetworks to generate tailored parameters for each layer. This dual design ensures an optimal balance between performance and parameter efficiency while introducing less than 0.05\% additional parameters, making it significantly more compact than existing methods.
Extensive experiments on diverse downstream tasks, few-shot settings and domain generalization demonstrate that SaS significantly enhances performance while maintaining superior parameter efficiency compared to existing methods, highlighting the importance of capturing both shared and layer-specific information in transfer learning. Code and data are available at \url{https://anonymous.4open.science/r/SaS-PETL-3565}.

\end{abstract} 

\section{Introduction}






The rapid growth of foundation models, particularly Transformers \cite{vaswani2017attention_transformer} based architecture, has fundamentally transformed the field of artificial intelligence. Pre-trained on massive datasets, these models have demonstrated extraordinary capabilities to learn rich, contextualized representations has led to state-of-the-art performance across a wide range of applications from natural language processing \citep{devlin2019_bert, brown2020language, wang2018glue, yang2019xlnet} to computer vision \cite{dosovitskiy2020image, touvron2021training, liu2021swin, carion2020end, zheng2021rethinking}.

However, fine-tuning these foundation models for specific downstream tasks presents significant challenges. With millions or even billions of parameters, fully fine-tuning the entire network demands extensive computational and memory resources. Moreover, maintaining a separate model weights for each downstream task becomes impractical as the number of tasks increases. Additionally, adapting such massive models to specialized tasks often requires careful adjustments to avoid overfitting, particularly when task-specific data is limited. These challenges highlight the need for innovative approaches that can harness the power of foundation models while reducing the fine-tuning overhead.

To address these challenges, parameter-efficient transfer learning (PETL) methods adapt large-scale models to new tasks by fine-tuning only a small subset of parameters or by incorporating lightweight auxiliary modules. This approach enhances both performance and computational efficiency by maintaining the major part of the pre-trained weights frozen. Pioneering methods in this domain such as Adapter \cite{houlsby2019parameter_adapter}, prompt tuning \cite{jia2022visual_vpt}, LoRA \cite{hu2022_lora}, SCT \cite{zhao2024_sct}, GPS \cite{zhang2024gradient_gps}, BitFit \cite{zaken2021_bitfit} have demonstrated promising results. 

We observe that most existing techniques process features at different layers using dedicated modules, each designed to capture task-specific information at its respective layer. However, while each layer retains unique information, they also share common features. This redundancy leads to the inclusion of excess parameters to capture the same shared information across multiple layers. We propose separating shared and layer-specific information into distinct modules, with a shared module to learn the common statistical characteristics. This approach significantly reduces the number of parameters while preserving effective feature extraction. Furthermore, a single shared module can be used across multiple layers of a foundation model, enabling it to capture and reinforce the common statistical characteristics of latent representations across layers. (see Section \ref{sec:method} for the detailed discussion).



To this end, in this paper, our approach leverages both the common statistical properties inherent in the latent representations across layers and the distinct features specific to each layer. We achieve this by integrating a shared module that operates across all layers with layer-specific modules that fine-tune features at each abstraction level. Aligned with our approach, weight-sharing techniques have been extensively explored and proven effective in natural language transformer models \cite{bai2019deep, dehghani2018universal, lan2019albert}, and also been partially adopted in lightweight Vision Transformer architectures, like MiniViT \cite{zhang2022minivit}, underscoring the feasibility of our idea in enhancing parameter efficiency and performance. The key contributions of this paper are summarized as follows:
\begin{itemize}
    \item In Section \ref{sec:motivations}, we analyze the distributional shift that occurs during the transfer learning process and examine how adjusting this shift with a small number of additional parameters can improve performance. We also highlight the importance of capturing both the common underlying information across layers and the unique, layer-specific features at each level of abstraction. 
    
    \item Building on this analysis, in Section \ref{sec:architecture}, we propose a simple and lightweight SaS approach that introduces less than 0.05\% additional parameters. Our method consists of two key components: a shared module that captures and reinforces the common statistical characteristics of latent representations across layers and layer-specific modules that capture the distinctive features learned at each stage of the model.

    \item Extensive experiments on various downstream tasks presented in Section \ref{sec:experiment} demonstrate that our SaS approach not only enhances performance but also balances parameter efficiency, validating its efficiency and robustness in multiple practical transfer learning scenarios.

\end{itemize}

\section{Related Works}
\label{sec:related_work}

Parameter-efficient transfer learning (PETL) has gained significant attention for adapting large pre-trained models to downstream tasks while minimizing computational overhead. Various methods have been proposed to achieve this goal.

\paragraph{Adapter Tuning.} Adapter \cite{houlsby2019parameter_adapter} is a pioneer method that introduces a bottleneck module within each Transformer layer, at both the Multi-Head Attention (MHA) and Feed-Forward Networks (FFN) layers. This module focuses on fine-tuning a small subset of parameters, reducing the total number of parameters that need to be trained. AdaptFormer \cite{chen2022_adaptformer} extends this idea by adding the adapter module in parallel to the FFN within each block in a pre-trained ViT. It also incorporates a non-linear activation function and a scaling factor to control the output. The SNF \cite{wang2023adapting_snf} approach incorporates Normalizing Flows modules into the residual connections of each block in ViT. This allows the model to capture complex data distributions during fine-tuning.

\paragraph{Prompt Tuning and Prefix Tuning.} VPT \cite{jia2022visual_vpt} introduces visual prompts to vision transformers for efficient adaptation. In VPT-Shallow \cite{jia2022visual_vpt}, the prompt is added to the input sequence before the first encoder layer, whereas VPT-Deep \cite{jia2022visual_vpt} appends the prompt to each encoder layer’s input. This approach allows for task-specific adaptation without altering the original model parameters. Similar to VPT, Prefix Tuning \cite{li2021_prefixtuning} uses trainable prefix tokens in each layer of the Transformer model. These trainable tokens act as task-specific adjustments while keeping the original parameters frozen, ensuring that the model can efficiently transfer to new tasks without the need for full-scale fine-tuning.

\paragraph{Side Tuning.} LST \cite{sung2022_lst} introduces a small and separate ladder-side network on the side of the pre-trained network. It takes intermediate activations from the backbone to process and make the prediction. DTL \cite{fu2024dtl} ensures the simplicity and effectiveness when fine-tuning large pre-trained network to downstream task with limited data by designing a Compact Side Network (CSN) within each block of ViT to extract task-specific information. These methods enhance task-specific learning by using additional networks that operate in parallel to the pre-trained model.


\paragraph{Specification Tuning and Reparameterization Tuning.} Besides increasing both parameters and computational cost during training and inference, specification tuning and reparameterization tuning approaches aim to fine-tune models by introducing additional matrices only during training while integrating them into the model’s weights during inference. This strategy minimizes computational overhead at inference time while still allowing effective task adaptation. 
In this category, BitFit \cite{zaken2021_bitfit} is a parameter-efficient method that fine-tunes only the bias terms of a pre-trained model while keeping all other parameters fixed. This approach introduces minimal modifications while still enabling task-specific adaptation. GPS \cite{zhang2024gradient_gps} adopts a gradient-based strategy to fine-tune a small, crucial subset of parameters (a sub-network), allowing for efficient adaptation without requiring modifications to the entire model.
SCT (Salient Channel Tuning) \cite{zhao2024_sct} focuses on optimizing a specific subset of salient channels by introducing Salient Channel Tuning Modules (SCTM), which are inserted after MHSA or FFN. This method ensures targeted fine-tuning, enhancing model efficiency.
LoRA (Low-Rank Adaptation) \cite{hu2022_lora} introduces a low-rank decomposition approach, integrating tunable pairs of low-rank weight matrices into each encoder layer of a pre-trained ViT. Similarly, FacT \cite{jie2023facT} adopts factorized weight matrices to achieve efficient parameter tuning while preserving performance.


\section{Methodology}
\label{sec:method}
\subsection{Motivations} \label{sec:motivations}

Foundation models are typically trained on massive datasets, allowing them to learn data distributions that are broad and generalized enough to cover a wide range of downstream tasks. These models capture complex patterns and features that are often common across various domains. However, when transferring learning from these pre-trained foundation models to a downstream task, the distribution of the downstream task may not exactly match the distribution learned during pre-training. Therefore, the goal is to adjust the downstream data distribution to align more closely with the distribution of the large-scale dataset that the foundation model was trained on.
To achieve this, one can equip each layer of a foundation model with an individual lightweight component, such as Adapter \cite{houlsby2019parameter_adapter}, prompt tuning \cite{jia2022visual_vpt}, or LoRA \cite{hu2022_lora}, and fine-tune these additional components to adapt to new data. This strategy helps mitigate the shift between the massive dataset used to train foundation models and the new dataset, thereby improving performance on the latter. However, while each lightweight component introduces only a small number of parameters, their accumulation across all layers may over-parameterize the new dataset, especially if it is small scale, potentially leading to overfitting and performance degradation. 

A common approach to drastically reducing the number of parameters in additional components is to apply the parameter-sharing principle. Specifically, a single shared module can be used across multiple layers of a foundation model, enabling it to capture and reinforce the common statistical characteristics of latent representations across layers. 
To demonstrate the effectiveness of this approach, we conduct experiments on datasets where a single module is shared across the layers of a ViT. The results show a significant improvement over the linear probing predictions of foundation models in Table \ref{tab:ablation_AB_bias}, highlighting the shared module's ability to effectively exploit the common statistical structure of latent representations across layers.   


\begin{table}[]
\centering

\resizebox{1.0\columnwidth}{!}
{\begin{tabular}{ccccccc}
\toprule
Method & \#p & Nat. & Spe. & Str. & Avg. \\ \midrule \midrule

Linear probing \cite{jia2022visual_vpt} & 0M & 68.92 & 77.15 & 25.6 & 52.94 \\
$F_{sh}(z^i)$ & 0.019M & 82.30 & 85.66 & 61.75 & 74.40  \\
$F_{sh}(z^i) + b^i$ & 0.046M & 82.11 &  86.12 & 62.57 & 74.90 \\
\bottomrule
\end{tabular}}
\caption{Comparison of fine-tuning ViT-B/16 pre-trained model on ImageNet-21k with different approach. $F_{sh}(\cdot)$ is a lightweight module with low-rank projections to share across layers to capture common statistics. $b^i$ is a bias vector added to the layer $i$ to present for layer-specific information. $z^i$ is input of layer $i$ in model.}
\label{tab:ablation_AB_bias}
\end{table}

\label{sec:ablation_study}

However, despite similarities in data distributions across layers or blocks, intrinsic differences remain in how each layer processes information. For instance, earlier layers primarily capture low-level features such as edges and textures, while deeper layers extract more complex, high-level representations. To account for these varying statistical characteristics across layers, we propose incorporating layer-specific parameters into the shared module. Moreover, these layer-specific parameters must be designed in an extremely economical manner. To demonstrate its effectiveness, we conduct experiments where a bias vector $b$ is added to each layer as a lightweight modification. The experimental results in Table \ref{tab:ablation_AB_bias} highlight the advantage of incorporating layer-specific information in adapting to the distinct statistical properties of different layers. 

Although adding a simple and economical bias vector enhances performance, it lacks the sophistication needed to fully capture the inherent complexity of statistical variations across layers. This raises a crucial question: \textit{How can we design layer-specific parameters that are both lightweight and powerful enough to effectively handle these complexities?}  

To address this, we propose a highly compact hyper-network to generate layer-specific parameters. Specifically, various lightweight, learnable layer-specific inputs are fed into a shared hyper-network, which then produces the required parameters for each layer. This approach provides an efficient mechanism to generate refined and adaptive layer-specific parameters, enabling the model to effectively capture the distinctive features learned at each stage.      



\subsection{Specific and shared modules}
\label{sec:architecture}

Given a foundation backbone containing L blocks and an input $x$, the output $z^L$ of this foundation model is calculated as $z = B^L(B^{L-1}(...B^0(x))$ with $z^{i+1} = B^i(z^{i})$ is intermediate feature of each block $i$ and $z^1 = x$.
 
\begin{figure*}[h!]
     
     \begin{subfigure}[b]{0.6\textwidth}
         \includegraphics[width=\linewidth]{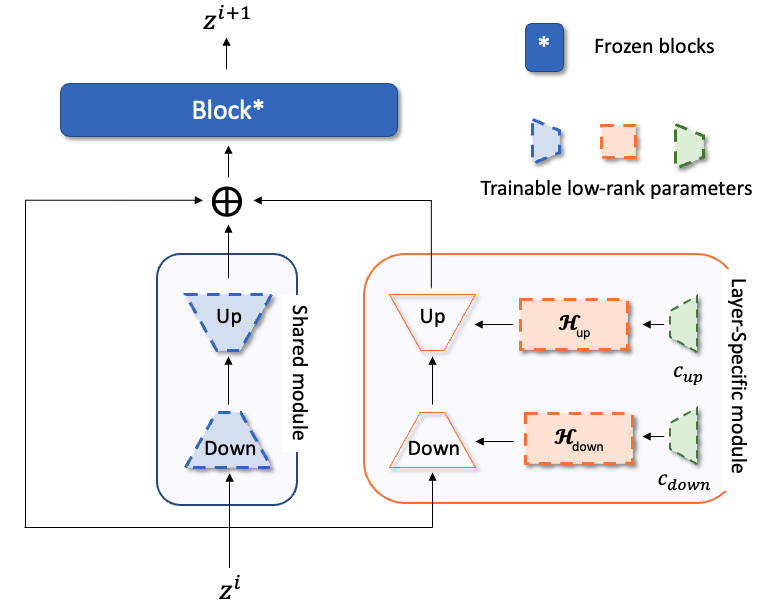}
     \end{subfigure}
     \centering
     
    \caption{Proposed architecture of SaS that consists of two modules: the shared module employs low-rank projection techniques to capture and reinforce the common statistical characteristics present in the dataset; Layer-specific module a Hyper-network to generate parameters for each individual layer, tailoring features at each abstract level. }
    \label{fig:module}
     
\end{figure*}

We design our module to adjust the input $z^l$ of each block as:
\vspace{-3mm}
\begin{align*}
    & \Tilde{z}^i = F_{sh}\left(z^i\right) + G^i\left(z^i\right) \\
    & z^{i+1} = B^i\left(z^i + \Tilde{z}^i \right)
\end{align*}
where  $F_{sh}$ is the shared module across layers and $G^i$ is a layer-specific module, generated by a hyper-network. The overall proposed architecture is presented in Figure \ref{fig:module}.

\paragraph{Shared module.} We employ a bottleneck architecture that includes a down-projection layer $W_{up} \in \mathbb{R}^{d' \times d}$ and an up-projection layer $W_{down} \in \mathbb{R}^{d' \times d}$. This design helps maintain computational efficiency by reducing the number of parameters required for the shared module. The calculation for this module is given by:
\begin{align*}
    F_{sh}(z^i) = \sigma\left(z^i\left(W_{down}\right)^TW_{up}\right)
\end{align*}
where $\sigma(\cdot)$ represents an activation function and $W_{down}$ and $W_{up}$ are shared across layers of foundation models. 

This bottleneck structure effectively reduces the dimensionality of the intermediate features as we set $d' << d$, allowing for a more compact representation and optimizing parameter efficiency, while still enabling the model to learn essential patterns and transformations for downstream tasks.


\paragraph{Layer-specific modules.} To improve parameter efficiency, we dedicate a hyper-network \cite{ha2016hypernetworks} with various layer-specific inputs to generate layer-specific modules at different layers. We now discuss how to design our hyper-network.  


To be specific, we formulate our hyper-network at layer \( i \) using low-rank matrices \( \mathcal{H}_{up} \) and \( \mathcal{H}_{down} \), along with layer-specific inputs \( c^i_{up} \) and \( c^i_{down} \). This is detailed as follows:
\begin{align*}
    & W^i_{up} = c^i_{up}\mathcal{H}_{up} \\
    & W^i_{down} = c^i_{down}\mathcal{H}_{down} \\
    & G^i(z^i) = z^i\left(W^i_{down}\right)^TW^i_{up} 
\end{align*}
where $\mathcal{H}_{up}, \mathcal{H}_{down} \in \mathbb{R}^{r \times d} $ are trainable parameters of the hyper-network, $c^i_{up}$,  $c^i_{down} \in \mathbb{R}^{r' \times r}$ are trainable inputs of the hyper-network. Moreover, we set $r, r' << d$ to reduce the number of parameters.

Ideally, a single hyper-network would be used for all layers of a foundation model. However, to better capture layer-specific features, we employ \( M \) hyper-networks, with each one covering \( L/M \) layers. This approach balances parameter efficiency with the model’s ability to capture distinct features at different depths, enhancing both performance and resource utilization.



\subsection{Complexity Analysis}




The shared module consists of two matrices, each with dimensions $d' \times d$, resulting in a total of $2 d' d$ parameters. Since this module is shared across all layers (L layers in total), the total number of parameters introduced by the Shared module is $2 d' d$, independent of the number of layers. This design enables us to increase the dimensionality ($d'$) without significantly increasing the overall parameter count, thus preserving parameter efficiency.

The layer-specific module comprises $M$  hyper-networks, each containing two projection matrices of dimensions $r \times d$, resulting in a total of $2 M r d$ parameters. Additionally, each layer includes unique, trainable inputs for the hyper-network, denoted as $c^i_{up}$ and $c^i_{down}$, each of size $r' \times r$. These inputs add a total of $2 L r' r$ parameters across all layers.

Therefore, the total number of parameters introduced by our method is $2\left(d'd + Mrd + Lr'r\right)$.
Since we set $d', r', r << d$, our approach maintains a high level of parameter efficiency. Here, $L$ denotes the total number of blocks (layers) within the network, and $M$ represents the number of Layer-specific module pairs shared across the model. Notably, even as the pre-trained model becomes deeper (with an increased number of blocks, $L$), the total parameter count remains minimally affected due to the small magnitude of the product $r' r$. This design ensures scalability and parameter efficiency, particularly for large-scale and deep architectures.

Additionally, increasing the number of layer-specific modules $M$ does not significantly affect computational cost because it only depends on the size of projection matrices and input of hyper-networks, which remains constant regardless of how many layer-specific modules $M$ are employed. Thus, our method effectively scales with more layer-specific modules without increasing computational complexity.

\section{Experiments}
\begin{table*}[h!]
\centering
\resizebox{2.0\columnwidth}{!}
{\begin{tabular}{l|ccccccc|cccc|cccccccc|ccc}
\toprule
\multirow{2}{*}{Method} & \multicolumn{7}{c|}{\textbf{Natural}}            & \multicolumn{4}{c|}{\textbf{Specialized}} & \multicolumn{8}{c|}{\textbf{Structured}} &  &  &  \\
  &  \rot{CIFAR-100} &  \rot{Caltech101} &  \rot{DTD} &  \rot{Flowers102} &  \rot{Pets} &  \rot{SVHN} &  \rot{Sun397} &  \rot{Patch Camelyon} &  \rot{EuroSAT} &  \rot{Resisc45} &  \rot{Retinopathy} &  \rot{Clevr\/count} &  \rot{Clevr\/distance} &  \rot{DMLab} &  \rot{KITTI\/distance} &  \rot{dSprites\/loc} &  \rot{dSprites/ori} &  \rot{SmallNORB\/azi} &  \rot{SmallNORB\/ele} & \rot{Mean} &  \rot{\#Params (M)} &  \rot{PPT} \\ \midrule \midrule


\multicolumn{20}{l}{\multirow{2}{*}{\textbf{Traditional Finetuning}}  }  &   &  &  \\    
\multicolumn{20}{l}{\multirow{2}{*}{}  }   &   &  & \\
Full fine-tuning \cite{jia2022visual_vpt} &  68.9 & 87.7 & 64.3 & 97.2 & 86.9 & 87.4 & 38.8 & 79.7 & 95.7 & 84.2 & 73.9 & 56.3 & 58.6 & 41.7 & 65.5 & 57.5 & 46.7 & 25.7 & 29.1 & 65.57 & 85.8M & - \\ 
Linear probing \cite{jia2022visual_vpt} &  63.4 & 85.0 & 63.2 & 97.0 & 86.3 & 36.6 & 51.0 & 78.5 & 87.5 & 68.6 & 74.0 & 34.3 & 30.6 & 33.2 & 55.4 & 12.5 & 20.0 & 9.6 & 19.2 & 52.94 & 0M & 0.53 \\  \midrule \midrule


\multicolumn{20}{l}{\multirow{2}{*}{\textbf{PETL Algorithms}}  }    &   &  &  \\
\multicolumn{20}{l}{\multirow{2}{*}{}  }  &   &  &  \\
Adapter \cite{houlsby2019parameter_adapter}  & 69.2  &  90.1  &  68.0  &  98.8  &  89.9  &  82.8  &  54.3  &  84.0  &  94.9  &  81.9  &  75.5  &  80.9  &  65.3  &  48.6  &  78.3  &  74.8  &  48.5  &  29.9  &  41.6  &  71.44 & 0.16M & 0.709 \\
VPT-Shallow \cite{jia2022visual_vpt} & 77.7  &  86.9  &  62.6  &  97.5  &  87.3  &  74.5  &  51.2  &  78.2  &  92.0  &  75.6  &  72.9  &  50.5  &  58.6  &  40.5  &  67.1  &  68.7  &  36.1  &  20.2  &  34.1  &  64.85  &  0.08M & 0.646
  \\
VPT-Deep \cite{jia2022visual_vpt} & 78.8  &  90.8  &  65.8  &  98.0  &  88.3  &  78.1  &  49.6  &  81.8  &  96.1  &  83.4  &  68.4  &  68.5  &  60.0  &  46.5  &  72.8  &  73.6  &  47.9  &  32.9  &  37.8  &  69.43    &  0.56M & 0.678 \\
BitFit \cite{zaken2021_bitfit} & 72.8  &  87.0  &  59.2  &  97.5  &  85.3  &  59.9  &  51.4  &  78.7  &  91.6  &  72.9  &  69.8  &  61.5  &  55.6  &  32.4  &  55.9  &  66.6  &  40.0  &  15.7  &  25.1  &  62.05    &  0.10M & 0.617\\
LoRA \cite{hu2022_lora} & 67.1  &  91.4  &  69.4  &  98.8  &  90.4  &  85.3  &  54.0  &  84.9  &  95.3  &  84.4  &  73.6  &  82.9  &  69.2  &  49.8  &  78.5  &  75.7  &  47.1  &  31.0  &  44.0  &  72.25   &  0.29M & 0.713  \\
AdaptFormer \cite{chen2022_adaptformer} & 70.8  &  91.2  &  70.5  &  99.1  &  90.9  &  86.6  &  54.8  &  83.0  &  95.8  &  84.4  &  76.3  &  81.9  &  64.3  &  49.3  &  80.3  &  76.3  &  45.7  &  31.7  &  41.1  &  72.32    &  0.16M & 0.718 \\
SSF \cite{lian2022scaling_ssf} & 69.0  &  92.6  &  75.1  &  99.4  &  91.8  &  90.2  &  52.9  &  87.4  &  95.9  &  87.4  &  75.5  &  75.9  &  62.3  &  53.3  &  80.6  &  77.3  &  54.9  &  29.5  &  37.9  &  73.10   &  0.21M & 0.724  \\
NOAH \cite{zhang2024neural_noah}  & 69.6  &  92.7  &  70.2  &  99.1  &  90.4  &  86.1  &  53.7  &  84.4  &  95.4  &  83.9  &  75.8  &  82.8  &  68.9  &  49.9  &  81.7  &  81.8  &  48.3  &  32.8  &  44.2  &  73.25    &  0.43M & 0.719 \\
SCT \cite{zhao2024_sct}  & 75.3  &  91.6  &  72.2  &  99.2  &  91.1  &  91.2  &  55.0  &  85.0  &  96.1  &  86.3  &  76.2  &  81.5  &  65.1  &  51.7  &  80.2  &  75.4  &  46.2  &  33.2  &  45.7  &  73.59    &  0.11M & 0.732 \\
FacT \cite{jie2023facT}  & 70.6  &  90.6  &  70.8  &  99.1  &  90.7  &  88.6  &  54.1  &  84.8  &  96.2  &  84.5  &  75.7  &  82.6  &  68.2  &  49.8  &  80.7  &  80.8  &  47.4  &  33.2  &  43.0  &  73.23   &  0.07M & 0.730 \\
RepAdapter \cite{luo2023towards_RepAdapter}  & 72.4  &  91.6  &  71.0  &  99.2  &  91.4  &  90.7  &  55.1  &  85.3  &  95.9  &  84.6  &  75.9  &  82.3  &  68.0  &  50.4  &  79.9  &  80.4  &  49.2  &  38.6  &  41.0  &  73.84   &  0.22M & 0.731 \\
Hydra \cite{kim2024_hydra}  & 72.7  &  91.3  &  72.0  &  99.2  &  91.4  &  90.7  &  55.5  &  85.8  &  96.0  &  86.1  &  75.9  &  83.2  &  68.2  &  50.9  &  82.3  &  80.3  &  50.8  &  34.5  &  43.1  &  74.21   &  0.28M & 0.733 \\

LST \cite{sung2022_lst}  & 59.5  &  91.5  &  69.0  &  99.2  &  89.9  &  79.5  &  54.6  &  86.9  &  95.9  &  85.3  &  74.1  &  81.8  &  61.8  &  52.2  &  81.0  &  71.7  &  49.5  &  33.7  &  45.2  &  71.70   &  2.38M & 0.653 \\
DTL \cite{fu2024dtl}  & 69.6  &  94.8  &  71.3  &  99.3  &  91.3  &  83.3  &  56.2  &  87.1  &  96.2  &  86.1  &  75.0  &  82.8  &  64.2  &  48.8  &  81.9  &  93.9  &  53.9  &  34.2  &  47.1  &  74.58  &  0.04M & 0.744 \\
HST \cite{lin2023hierarchical_hst}  & 76.7  &  94.1  &  74.8  &  99.6  &  91.1  &  91.2  &  52.3  &  87.1  &  96.3  &  88.6  &  76.5  &  85.4  &  63.7  &  52.9  &  81.7  &  87.2  &  56.8  &  35.8  &  52.1  &  \textbf{75.99}   &  0.78M & 0.735 \\ 
GPS \cite{zhang2024gradient_gps}  & 81.1  &  94.2  &  75.8  &  99.4  &  91.7  &  91.6  &  52.4  &  87.9  &  96.2  &  86.5  &  76.5  &  79.9  &  62.6  &  55.0  &  82.4  &  84.0  &  55.4  &  29.7  &  46.1  &  75.18   &  0.22M & 0.744 \\
 LAST \cite{tang2024low_last} & 66.7  &  93.4  &  \textbf{76.1}  &  \textbf{99.6}  &  89.8  &  86.1  &  54.3  &  86.2  &  96.3  &  86.8  &  75.4  &  81.9  &  65.9  &  49.4  &  82.6  &  87.9  &  46.7  &  32.3  &  51.5  &  74.15  &  0.66M & 0.721 \\
 SNF \cite{wang2023adapting_snf} & \textbf{84.0}  &  94.0  &  72.7  &  99.3  &  91.3  &  90.3  &  54.9  &  87.2  &  97.3  &  85.5  &  74.5  &  82.3  &  63.8  &  49.8  &  82.5  &  75.8  &  49.2  &  31.4  &  42.1  &  74.10  &  0.25M & 0.733 \\ \midrule
 SaS (M=4) & 75.9 & 94.1 & 71.3 & 99.3 & 91.0 & 91.7 & 55.4 & 86.8 & 96.2 & 87.8 & 73.6 & 83.9 & 63.4 & 50.9 & 81.2 & 85.6 & 55.5 & 33.4 & 45.9 & 74.89 &  0.04M & \textbf{0.747} \\
 SaS (M=6) & 75.6 & \textbf{95.0} & 70.7 & 99.3 & 91.0 &  91.5 & 55.9 & 87.8 & 95.7 & \textbf{87.1} & 73.5 & \textbf{83.6} & 64.6 & 50.0 & 80.7& \textbf{89.0} & 55.0 & 34.6 & 48.1 & 75.20  & 0.05M & \textbf{0.750}\\ \bottomrule
\end{tabular}
}
\caption{Benchmark results on VTAB with ViT-B/16 models pre-trained on ImageNet-21K. Each experiment is conducted with three different random seeds and reports an average score.}
\label{tab:vit_vtab}
\end{table*}

In this section, we conduct extensive experiments across various settings to validate the effectiveness of our SaS model:

\begin{itemize}
    \item \textit{Image classification Benchmark}: We evaluate SaS with different backbone architectures on the VTAB-1k \cite{zhai2019large_vtab} and FGVC \cite{wah2011caltech_cub_dataset, van2015building_nabird_dataset, nilsback2008automated_flower_dataset, dataset2011novel_standford_dog_dataset, gebru2017fine_standford_car_dataset} benchmarks, testing multiple variations of our approach.
    \item  \textit{Domain generalization setting}: We assess the robustness of SaS in realistic scenarios with distribution shifts problem using the ImageNet dataset and its derived variants. 
    \item \textit{Few-shot Learning}: To further analyze the impact of SaS under data-limited conditions, we explore few-shot learning across five diverse datasets, systematically varying the number of training shots per class. 
\end{itemize}
The details of each setting, including the setup and results, are presented in the following sections.

\subsection{Experimental Settings}

\textbf{Pre-trained backbone.} To ensure a fair comparison, we follow the V-PETL Bench \cite{xin2024v_v_petf} methodology and primarily use the ViT-B/16 model, pre-trained on ImageNet-21K, as the initialization for fine-tuning. These architectures are commonly used in the visual PETL domain. Additionally, we extend our method to the Swin Transformer with the Swin-B backbone, which utilizes a hierarchical transformer-based architecture. The pre-trained weights for both the ViT and Swin Transformers are publicly available.

\textbf{Initialization.} We ensure that both the Share module and Layer-specific module produce either zero or sufficiently small values at the early stages of fine-tuning, preventing the model from deviating too far from the original pre-trained weights. Specifically, for the Share module, we initialize the down-projection $W_{down}$ using Kaiming Normal \cite{he2015delving} and set the up-projection $W_{up}$ to zero. For the Layer-specific module, we initialize the weight matrices $\mathcal{H}_{down}$ using Kaiming Normal, $\mathcal{H}_{up}$ with zeros, and the inputs $c^i_{up}$ and $c^i_{down}$ with small values of $1e-5$. This strategy helps maintain the model's stability while allowing for a gradual adaptation to new downstream tasks.

\subsection{Performance on Image Classification}

\textbf{VTAB-1k dataset} \cite{zhai2019large_vtab} consists of 19 distinct datasets, which are grouped into three categories: Natural, Specialized, and Structured. Each dataset contains only 1,000 images for training, making the task challenging due to the limited amount of data. Additionally, the images show significant variation in data distribution across the datasets, further complicating the learning process.

We evaluate our proposed architecture on the VTAB-1k dataset using both ViT-B/16 and Swin-B backbones. For ViT-B/16, we test two variations of SaS with $M=4$ and $M=6$. The Swin-B model is designed with four layers with number of block for each layer is [2,2, 18, 2], the embed dimension for each layer is [128, 256, 512, 1024]. So that, we need use a different Share module and $M$ Hypernets for each layer. Pariticularly, we use $M=[1,1,6,1]$ for corresponding layer in Swin-B architecture. For all experiments, we set $d' = r' = 8$ and $r = 4$. 

In Table \ref{tab:vit_vtab}, our SaS achieves comparable performance with significantly fewer parameters. With the same 0.04M parameters, SaS achieves 0.31\% higher average accuracy than DTL \cite{fu2024dtl}, resulting in a 0.747 PPT score. By slightly increasing the number of parameters for M=6, SaS reaches an average accuracy of 75.2\%, making it the runner-up compared to the current state-of-the-art method HST, which uses $15.6\times$ more parameters. Overall, SaS achieves the highest PPT score compared to the baselines.

In Table \ref{tab:swin_vtab}, SaS outperforms DTL \cite{fu2024dtl} in both accuracy and PPT score, despite using fewer parameters. While the overall score metrics are slightly lower compared to other DTL \cite{fu2024dtl} variations, SaS stands out by achieving the highest accuracy on the Natural group of the VTAB-1k dataset, highlighting its strong performance in this category.

\begin{table}[h!]
\centering
\begin{tabular}{ccccccc}
\toprule
Method & \#p & Nat. & Spe. & Str. & Avg.  & PPT \\ \midrule \midrule

Full \cite{jia2022visual_vpt} & 86.7 & 79.2 & 86.2 & 59.7 & 75.0 &  - \\
Linear \cite{jia2022visual_vpt} & 0 & 73.5 & 80.8 & 33.5 & 62.6 &  0.626 \\
BitFit \cite{zaken2021_bitfit} & 0.20 & 74.2 & 80.1 & 42.4 & 65.6 &  0.650 \\
VPT \cite{jia2022visual_vpt} & 0.16  & 76.8 & 84.5 & 53.4 & 71.6 &  0.711 \\
FacT \cite{jie2023facT} & 0.14 & 83.1 & 86.9 & 62.1 & 77.4 & 0.769  \\
DTL \cite{fu2024dtl} & 0.09  & 82.4 & 87.0 & 64.2 & 77.9 &  0.776  \\
DTL+ \cite{fu2024dtl}  & 0.13   & 82.4 & 86.8 & \textbf{66.0} & 78.4 &  0.779  \\
DTL+* \cite{fu2024dtl} & 0.14  & 83.2 & \textbf{87.0} & 65.7 & 78.6 &  0.781 \\  \midrule
SaS  &  0.07 &  \textbf{83.7} &  86.2 &  63.9 & 78.0 &  0.778\\  \bottomrule
\end{tabular}
\caption{Results on VTAB-1K with Swin-B backbone. ‘\#p’ is the number of trainable parameters. Nat./Spe./Str./Avg. are the results in three VTAB-1k groups and their group-wise average.}
\label{tab:swin_vtab}
\end{table}

\textbf{FGVC dataset.} The FGVC benchmark comprises five fine-grained datasets for visual classification tasks: CUB-100-2011 \cite{wah2011caltech_cub_dataset}, NABirds \cite{van2015building_nabird_dataset}, Oxford Flowers \cite{nilsback2008automated_flower_dataset}, Stanford Dogs \cite{dataset2011novel_standford_dog_dataset},  and Stanford Cars \cite{gebru2017fine_standford_car_dataset}. Each dataset contains between 1,000 and 21,000 images for training, offering a diverse range of challenges for fine-grained image recognition. We follow the same setup for experiment with the VTAB-1k benchmark on ViT-B/16 to set $M=6, d'= 8, r' = 8, \text{and } r = 4$

As shown in Table \ref{tab:fgvc}, our SaS method achieves an impressive balance between efficiency and performance. Despite using significantly fewer parameters, it maintains the highest PPT score among all evaluated models, while its average accuracy remains on par with these baselines. it underscores the effectiveness of our approach in delivering both exceptional parameter efficiency and competitive performance.

\begin{table*}[h!]
\centering
\resizebox{1.7\columnwidth}{!}
{\begin{tabular}{lcccccccc}
\toprule
\multicolumn{1}{c}{\multirow{2}{*}{Method}} & CUB-200    &   \multirow{2}{*}{NABirds}  &   Oxford  &   Stanford &  Stanford & \multirow{2}{*}{Mean} & \multirow{2}{*}{\# Params (M)} & \multirow{2}{*}{PPT} \\
   & -2011 &  & Flowers & Dogs & Cars &  &  \\ \midrule \midrule
\multicolumn{9}{c}{\textbf{Traditional Finetuning}}                                                                                           \\
Full fine-tuning \cite{jia2022visual_vpt} & 87.3 & 82.7  &  98.8     &   89.4    &   84.5  &   88.54  &   85.8M & -  \\ 
Linear probing \cite{jia2022visual_vpt} &  85.3 & 75.9 & 97.9 & 86.2 & 51.3 & 79.32 & 0M & 0.793 \\ \midrule \midrule
\multicolumn{9}{c}{\textbf{PETL Algorithms}}                                                                                           \\
Adapter \cite{houlsby2019parameter_adapter} & 87.1 & 		84.3 & 98.5 & 89.8 & 		68.6  & 85.66 & 0.41M & 0.842 \\
AdaptFormer \cite{NEURIPS2022_adapformer} & 88.4 & 		84.7 & 99.2 & 88.2 & 		81.9 &  88.48 & 0.46M & 0.868 \\
Prefix Tuning \cite{li2021_prefixtuning}  & 87.5 & 		82.0 & 98.0 & 74.2 & 		90.2 & 86.38 & 0.36M & 0.851 \\
U-Tuning \cite{jiang2023rethinking_utuning}   & 89.2 & 		85.4 & 99.2 & 84.1 & 		92.1 &  90.00 & 0.36M & 0.886 \\
BitFit \cite{zaken2021_bitfit}  & 87.7 & 		85.2 & 99.2 & 86.5 & 		81.5 & 88.02 & 0.10M & 0.876 \\
VPT-Shallow \cite{jia2022visual_vpt} & 86.7 & 		78.8 & 98.4 & 90.7 & 		68.7  & 84.66 & 0.25M & 0.837 \\
VPT-Deep \cite{jia2022visual_vpt} & 88.5 & 		84.2 & 99.0 & 90.2 & 		83.6  &  89.10 & 0.85M & 0.859 \\
SSF \cite{lian2022scaling_ssf} & 89.5 & 		85.7 & 99.6 & 89.6 & 		89.2  &  90.72 & 0.39M & 0.892 \\
LoRA \cite{hu2022_lora} & 85.6 & 		79.8 & 98.9 & 87.6 & 		72.0  & 84.78 & 0.77M & 0.821 \\
GPS \cite{zhang2024gradient_gps} & 89.9 & 86.7 & 99.7 & 92.2 & 90.4 & \textbf{91.78} & 0.66M & 0.893 \\
HST \cite{lin2023hierarchical_hst}  & 89.2 & 		85.8 & 99.6 & 89.5 & 		88.2  &  90.46 & 0.78M & 0.876 \\
LAST \cite{tang2024low_last}    &  88.5  &  84.4 &   99.7 &  86.0  &  88.9  &  89.50 & 0.66M & 0.870  \\
SNF \cite{wang2023adapting_snf}     & 90.2  &  87.4 &  99.7 & 89.5  &  86.9 &  90.74 &  0.25M   & 0.898  \\
\midrule 
SaS (M=6)  &  89.2  & 85.7  &  99.4   &  90.9  &  86.3 &   90.30   &  \textbf{0.05M}   &  \textbf{0.901} \\ \bottomrule
\end{tabular}
}
\caption{Image classification experiments on FGCV benchmarks with ViT-B/16 pre-trained models} 
\label{tab:fgvc}
\end{table*}

\subsection{Experiments on Domain Generalization}

\begin{table}[t!]
\centering
\resizebox{1.0\columnwidth}{!}
{\begin{tabular}{cccccc}
\bottomrule
\multicolumn{1}{c}{\multirow{2}{*}{Method}} & Source   & \multicolumn{4}{c}{Target}    \\ \cmidrule{3-6} 
\multicolumn{1}{c}{}  & ImageNet & -Sketch & -V2   & -A    & -R    \\  \midrule \midrule
Adapter  \cite{houlsby2019parameter_adapter}             & 70.5     & 16.4    & 59.1  & 5.5   & 21.1\\
VPT  \cite{jia2022visual_vpt}                 & 70.5     & 18.3    & 58.0  & 4.6   & 23.2  \\
LoRA \cite{hu2022_lora}                 & 70.8     & 20.0    & 59.3  & 6.9   & 23.3  \\
NOAH \cite{zhang2024neural_noah}                 & 71.5     & 24.8    & 66.1  & 11.9  & 28.5  \\
DTL  \cite{fu2024dtl}                 & 78.3     & 35.4    & 67.8  & 14.0  & 34.4  \\
DTL+    \cite{fu2024dtl}                & 78.7     & 35.7    & 67.8  & 14.2  & 34.4  \\  \midrule
SaSP (M=6)                  & \textbf{79.0}     & \textbf{36.8}   & \textbf{68.1} & \textbf{15.5} & \textbf{34.7} \\  \bottomrule
\end{tabular}}
\caption{Top-1 accuracy on domain generalization experiments with ViT-B/16 models pre-trained on ImageNet-21K as the backbone}
\label{tab:domain_gen}
\end{table}

We follow the domain generalization experiments in NOAH \cite{zhang2024neural_noah} to evaluate the robustness of our method under realistic scenarios where domain shift \cite{zhou2022domain} is inevitable. In this setup, the model is initially fine-tuned using a subset of the ImageNet-1K \cite{deng2009imagenet} dataset, with 16 images (16-shot) per class. Subsequently, this model is evaluated on three standard subsets originating from ImageNet-1K: the original ImageNet-1K validation set, ImageNet-V2 \cite{recht2019imagenet_v2}, and ImageNet-Sketch \cite{wang2019learning_sketch}. Additionally, to further examine the robustness and domain generalization capabilities of our method, we evaluate it on two challenging datasets: ImageNet-A \cite{hendrycks2021natural_imagenet_A}, comprising adversarially filtered examples, and ImageNet-R \cite{hendrycks2021many_imagaget_R}, featuring diverse artistic renditions of ImageNet-1K images. 


We follow the same experimental setup for image classification on ViT-B/16, setting $M = 6, d' = 8, r' = 8$, and $r = 4$. All results are reported as the mean performance across three independent runs, each initialized with different random seeds. As illustrated in Table \ref{tab:domain_gen}, our SaS method consistently delivers state-of-the-art performance, achieving the highest accuracy across all evaluated tasks. Notably, SaS surpasses baselines by 1.2\% on ImageNet-Sketch and 1.3\% on the challenging ImageNet-A dataset, demonstrating its effectiveness in handling distributional shifts. Remarkably, despite maintaining a comparable number of trainable parameters, SaS outperforms strong competitors such as DTL and DTL+ \cite{fu2024dtl}, further validating its robustness and superior domain generalization capabilities. These results underscore the strength of SaS in adapting to real-world domain shifts, making it a highly effective and parameter-efficient fine-tuning approach.

\subsection{Experiments on Few-shot Learning}

\label{sec:experiment}
In previous sections, we evaluated our SaS method against various baselines under both data-limited and data-sufficient scenarios, observing that our model performs particularly effectively when training data is limited. In this section, we extend our analysis to a few-shot learning setting by increasing the number of training samples (shots) per class from 1, 2, 4, 8, to 16. Following the experiments established in FacT \cite{jie2023facT}, we evaluate on five fine-grained datasets: FGVC-Aircraft \cite{maji2013fine_aircraf}, Oxford-Pets \cite{parkhi2012cats_oxfoedpet}, Food-101 \cite{bossard2014food_food101}, Stanford Cars \cite{krause20133d_standfordcar}, and Oxford-Flowers102 \cite{nilsback2006visual_Oxford_Flowers102}. Each experiment is repeated across three random seeds, and the mean accuracy is reported.

We follow the same setup for experiments of image classification on ViT-B/16 to set $M=6, d'= 8, r' = 8$ and  $r = 4$. As presented in Figure \ref{fig:fewshot}, our SaS method demonstrates remarkable performance, significantly outperforming the strongest baseline (DTL+) on the FGVC-Aircraft and Stanford Cars datasets. Specifically, SaS achieves substantial improvements of at least 3\% on the 8-shot setting and 4\% on the 16-shot setting. Overall, SaS consistently surpasses all baseline methods by a clear margin, highlighting its effectiveness and superior capability in few-shot fine-grained visual classification tasks.

\section{Ablation study}

\begin{figure*}[t!]

     \begin{subfigure}[b]{1.0\textwidth}
         \includegraphics[width=\linewidth]{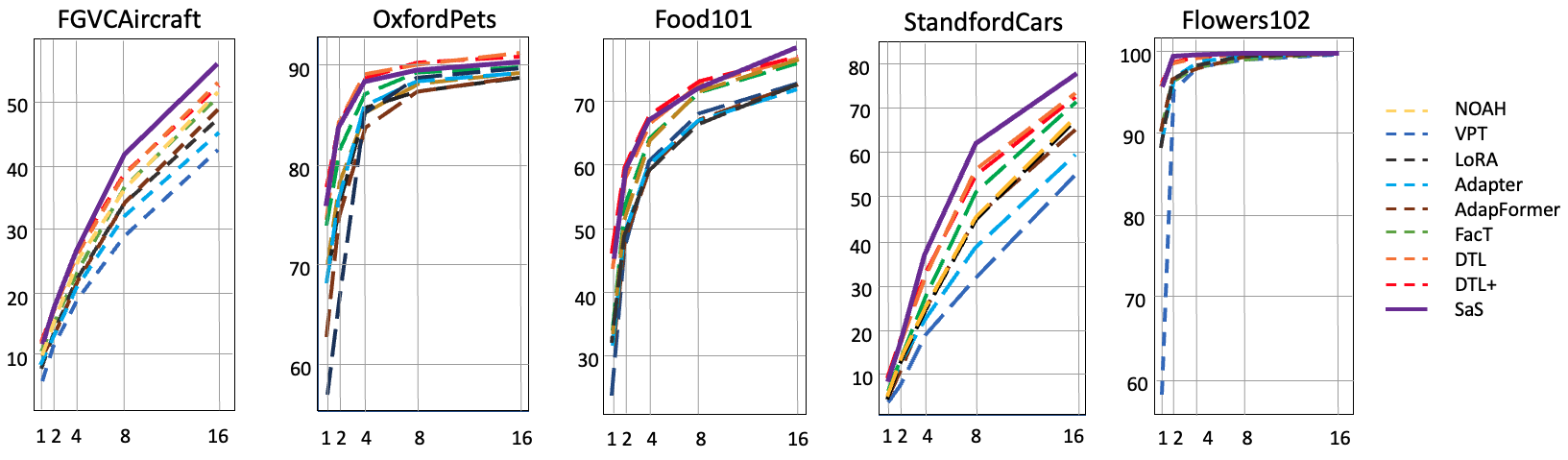}
         \label{fig:fewshot}
     \end{subfigure}
     \centering
    \caption{Top-1 accuracy on fine-grained few-shot benchmark with ViT-B/16 as the backbone. Best viewed in color}
    \label{fig:fewshot}
     
\end{figure*}
\subsection{Effectiveness of Shared module and Layer-specific module}

As discussed in Section \ref{sec:method}, our proposed method consists of two complementary modules: the Shared module, designed to efficiently capture the common features across all layers to facilitate general adaptation, and the Layer-specific module, which learns distinct representations tailored to each individual layer, enabling precise adjustments to the unique data distributions encountered at different depths within the network. This combination ensures a robust yet flexible fine-tuning process, improving task-specific performance while maintaining parameter efficiency. 

In this section, we analyze the contributions of each individual module in our proposed architecture on the CIFAR-100 dataset from the VTAB-1k benchmark. Specifically, we conduct experiments on three model variants. Model 1 consists of only Layer-specific module, configured with a low-rank setting of $r' = 8$ and $r = 4$. Model 2 employs one Shared module, configured with $d' = 8$. Model 3 integrates both modules, combining the Shared module with $d' = 8$ and the Layer-specific module with $r' = 4$ and $r = 2$. All three model variants have approximately 0.015M trainable parameters to ensure a fair comparison. The detailed results are presented in Table \ref{tab:ablation_share_spe}.

Our findings indicate that using only the Layer-specific module achieves competitive performance with an accuracy of 74.55\%. However, using only the Shared module notably surpasses this result, obtaining a higher accuracy of 75.40\%. This suggests that the Shared module effectively captures general knowledge presenting the same information from input across layers, significantly boosting performance even without layer-specific customization. Moreover, combining both modules further enhances accuracy to 75.60\%, demonstrating the complementary benefits of the general adaptation provided by the Shared module and the specialized adjustments from the Layer-specific module. This combined approach effectively balances generalizability and layer-specific refinement, leading to the best overall performance.

\begin{table}[h!]
\centering
\resizebox{1.0\columnwidth}{!}
{\begin{tabular}{cccc}
\toprule
Layer-Specific & \multirow{2}{*}{Shared module} &  \multirow{2}{*}{\# params} & \multirow{2}{*}{Accuracy} \\
Module & \\ \midrule \midrule

 $r'=16$, $r=8$ &- & 0.015M &  74.88  \\
 - & $d'=8$ & 0.015M & 75.40 \\ 
$r'=4$, $r=2$  & $d'=8$ & 0.015M & 75.60\\ \bottomrule
\end{tabular}}
\caption{Ablation study on effectiveness of Shared module and Layer-specific module}
\label{tab:ablation_share_spe}
\end{table}


\subsection{Effectiveness of $M$ Hypernets}

The Layer-specific module aims to capture the distinctive and unique information inherent to each individual layer, enabling the model to precisely adapt to the data distribution shifts that vary across layers. However, introducing completely separate projection parameters for each layer can significantly increase the number of trainable parameters.

To further enhance parameter efficiency while retaining the ability to learn layer-specific features, we adopt the Hypernet approach. Specifically, Hypernets are shared across multiple layers, but each layer is provided with a unique, learnable input, allowing the shared Hypernets to generate distinct projection weights for each layer. This strategy strikes a balance between parameter sharing and the specialization required. In this section, we analyze the impact of varying the number of Layer-specific modules ($M$) on model performance. Intuitively, using larger $M$ increases the total number of hyper-parameters, allowing the model to more flexibly capture the specific characteristics of each layer, potentially leading to improved accuracy. However, the overall computational cost remains constant despite varying $M$.

We conduct experiments on the VTAB-1k benchmark using the pre-trained ViT-B/16 network to analyze the impact of varying the number of Layer-specific modules (M) in our model. For each model variation, we consistently set hyperparameters as $d' = 8$, $r' = 8$, and $r = 4$. The detailed results are presented in Table \ref{tab:ablation_mHyper}. With the smallest setting (M=1), our SaS method already achieves remarkable performance, reaching an average accuracy of 74.30\%. Moreover, as we gradually increase the number of Hypernets (M), the overall accuracy consistently improves, demonstrating the effectiveness of employing multiple Hypernets to capture more specialized information across layers.

\begin{table}[h!]
\centering
\begin{tabular}{ccccccc}
\toprule
M & \#p & Nat. & Spe. & Str. & Avg.  & PPT \\ \midrule \midrule

1 & 0.019M & 82.1 & 85.6 & 61.9 & 74.30 &  0.742  \\
3 & 0.031M & 82.4 &  86.0 & 62.1 & 74.61 & 0.745 \\
4 & 0.037M & 82.7 & 86.1 & 62.5 & 74.89 & 0.748 \\
6 & 0.049M & 82.7 & 86.0 & 63.2 & 75.19 & 0.750  \\ \bottomrule
\end{tabular}
\caption{Effectiveness of varying the number of Hypernets (M) on model performance. The experiments are conducted on VTAB-1K with VIT-B/16 backbone.}
\label{tab:ablation_mHyper}
\end{table}

\vspace{-5mm}
\section{Conclusion}

In this work, we introduced SaS, a novel PETL approach that effectively adapts large-scale foundation models to downstream tasks while addressing distributional shifts in fine-tuning. SaS leverages a dual-module design that efficiently balances shared and layer-specific adaptation. Our approach consists of (1) a shared module, which captures common statistical characteristics across layers using low-rank projections, and (2) a layer-specific module, which employs hyper-networks to generate tailored parameters for each layer. This design allows SaS to achieve strong performance with less than 0.05\% additional parameters, making it significantly more compact than existing PETL methods. Through extensive experiments on diverse image classification, few-shot learning, and domain generalization, we demonstrate that SaS outperforms existing PETL approaches, highlighting the effectiveness of leveraging both shared representations and layer-specific adaptations to improve fine-tuning efficiency.



\clearpage

{
    \small
    \bibliographystyle{ieeenat_fullname}
    \bibliography{main}

\begin{thebibliography}{54}
\providecommand{\natexlab}[1]{#1}
\providecommand{\url}[1]{\texttt{#1}}
\expandafter\ifx\csname urlstyle\endcsname\relax
  \providecommand{\doi}[1]{doi: #1}\else
  \providecommand{\doi}{doi: \begingroup \urlstyle{rm}\Url}\fi

\bibitem[Bai et~al.(2019)Bai, Kolter, and Koltun]{bai2019deep}
Shaojie Bai, J~Zico Kolter, and Vladlen Koltun.
\newblock Deep equilibrium models.
\newblock \emph{Advances in neural information processing systems}, 32, 2019.

\bibitem[Bossard et~al.(2014)Bossard, Guillaumin, and Van~Gool]{bossard2014food_food101}
Lukas Bossard, Matthieu Guillaumin, and Luc Van~Gool.
\newblock Food-101--mining discriminative components with random forests.
\newblock In \emph{Computer vision--ECCV 2014: 13th European conference, zurich, Switzerland, September 6-12, 2014, proceedings, part VI 13}, pages 446--461. Springer, 2014.

\bibitem[Brown et~al.(2020)Brown, Mann, Ryder, Subbiah, Kaplan, Dhariwal, Neelakantan, Shyam, Sastry, Askell, et~al.]{brown2020language}
Tom Brown, Benjamin Mann, Nick Ryder, Melanie Subbiah, Jared~D Kaplan, Prafulla Dhariwal, Arvind Neelakantan, Pranav Shyam, Girish Sastry, Amanda Askell, et~al.
\newblock Language models are few-shot learners.
\newblock \emph{Advances in neural information processing systems}, 33:\penalty0 1877--1901, 2020.

\bibitem[Carion et~al.(2020)Carion, Massa, Synnaeve, Usunier, Kirillov, and Zagoruyko]{carion2020end}
Nicolas Carion, Francisco Massa, Gabriel Synnaeve, Nicolas Usunier, Alexander Kirillov, and Sergey Zagoruyko.
\newblock End-to-end object detection with transformers.
\newblock In \emph{European conference on computer vision}, pages 213--229. Springer, 2020.

\bibitem[Chen et~al.(2022{\natexlab{a}})Chen, GE, Tong, Wang, Song, Wang, and Luo]{NEURIPS2022_adapformer}
Shoufa Chen, Chongjian GE, Zhan Tong, Jiangliu Wang, Yibing Song, Jue Wang, and Ping Luo.
\newblock Adaptformer: Adapting vision transformers for scalable visual recognition.
\newblock In \emph{Advances in Neural Information Processing Systems}, pages 16664--16678. Curran Associates, Inc., 2022{\natexlab{a}}.

\bibitem[Chen et~al.(2022{\natexlab{b}})Chen, Ge, Tong, Wang, Song, Wang, and Luo]{chen2022_adaptformer}
Shoufa Chen, Chongjian Ge, Zhan Tong, Jiangliu Wang, Yibing Song, Jue Wang, and Ping Luo.
\newblock Adaptformer: Adapting vision transformers for scalable visual recognition.
\newblock \emph{Advances in Neural Information Processing Systems}, 35:\penalty0 16664--16678, 2022{\natexlab{b}}.

\bibitem[Dataset(2011)]{dataset2011novel_standford_dog_dataset}
E Dataset.
\newblock Novel datasets for fine-grained image categorization.
\newblock In \emph{First workshop on fine grained visual categorization, CVPR. Citeseer. Citeseer. Citeseer}, page~2. Citeseer, 2011.

\bibitem[Dehghani et~al.(2018)Dehghani, Gouws, Vinyals, Uszkoreit, and Kaiser]{dehghani2018universal}
Mostafa Dehghani, Stephan Gouws, Oriol Vinyals, Jakob Uszkoreit, and {\L}ukasz Kaiser.
\newblock Universal transformers.
\newblock \emph{arXiv preprint arXiv:1807.03819}, 2018.

\bibitem[Deng et~al.(2009)Deng, Dong, Socher, Li, Li, and Fei-Fei]{deng2009imagenet}
Jia Deng, Wei Dong, Richard Socher, Li-Jia Li, Kai Li, and Li Fei-Fei.
\newblock Imagenet: A large-scale hierarchical image database.
\newblock In \emph{2009 IEEE conference on computer vision and pattern recognition}, pages 248--255. Ieee, 2009.

\bibitem[Devlin et~al.(2019)Devlin, Chang, Lee, and Toutanova]{devlin2019_bert}
Jacob Devlin, Ming-Wei Chang, Kenton Lee, and Kristina Toutanova.
\newblock Bert: Pre-training of deep bidirectional transformers for language understanding.
\newblock In \emph{Proceedings of the 2019 conference of the North American chapter of the association for computational linguistics: human language technologies, volume 1 (long and short papers)}, pages 4171--4186, 2019.

\bibitem[Dosovitskiy et~al.(2020)Dosovitskiy, Beyer, Kolesnikov, Weissenborn, Zhai, Unterthiner, Dehghani, Minderer, Heigold, Gelly, et~al.]{dosovitskiy2020image}
Alexey Dosovitskiy, Lucas Beyer, Alexander Kolesnikov, Dirk Weissenborn, Xiaohua Zhai, Thomas Unterthiner, Mostafa Dehghani, Matthias Minderer, Georg Heigold, Sylvain Gelly, et~al.
\newblock An image is worth 16x16 words: Transformers for image recognition at scale.
\newblock \emph{arXiv preprint arXiv:2010.11929}, 2020.

\bibitem[Fu et~al.(2024)Fu, Zhu, and Wu]{fu2024dtl}
Minghao Fu, Ke Zhu, and Jianxin Wu.
\newblock Dtl: Disentangled transfer learning for visual recognition.
\newblock In \emph{Proceedings of the AAAI Conference on Artificial Intelligence}, pages 12082--12090, 2024.

\bibitem[Gebru et~al.(2017)Gebru, Krause, Wang, Chen, Deng, and Fei-Fei]{gebru2017fine_standford_car_dataset}
Timnit Gebru, Jonathan Krause, Yilun Wang, Duyun Chen, Jia Deng, and Li Fei-Fei.
\newblock Fine-grained car detection for visual census estimation.
\newblock In \emph{Proceedings of the AAAI Conference on Artificial Intelligence}, 2017.

\bibitem[Ha et~al.(2016)Ha, Dai, and Le]{ha2016hypernetworks}
David Ha, Andrew Dai, and Quoc~V Le.
\newblock Hypernetworks.
\newblock \emph{arXiv preprint arXiv:1609.09106}, 2016.

\bibitem[He et~al.(2015)He, Zhang, Ren, and Sun]{he2015delving}
Kaiming He, Xiangyu Zhang, Shaoqing Ren, and Jian Sun.
\newblock Delving deep into rectifiers: Surpassing human-level performance on imagenet classification.
\newblock In \emph{Proceedings of the IEEE international conference on computer vision}, pages 1026--1034, 2015.

\bibitem[Hendrycks et~al.(2021{\natexlab{a}})Hendrycks, Basart, Mu, Kadavath, Wang, Dorundo, Desai, Zhu, Parajuli, Guo, et~al.]{hendrycks2021many_imagaget_R}
Dan Hendrycks, Steven Basart, Norman Mu, Saurav Kadavath, Frank Wang, Evan Dorundo, Rahul Desai, Tyler Zhu, Samyak Parajuli, Mike Guo, et~al.
\newblock The many faces of robustness: A critical analysis of out-of-distribution generalization.
\newblock In \emph{Proceedings of the IEEE/CVF international conference on computer vision}, pages 8340--8349, 2021{\natexlab{a}}.

\bibitem[Hendrycks et~al.(2021{\natexlab{b}})Hendrycks, Zhao, Basart, Steinhardt, and Song]{hendrycks2021natural_imagenet_A}
Dan Hendrycks, Kevin Zhao, Steven Basart, Jacob Steinhardt, and Dawn Song.
\newblock Natural adversarial examples.
\newblock In \emph{Proceedings of the IEEE/CVF conference on computer vision and pattern recognition}, pages 15262--15271, 2021{\natexlab{b}}.

\bibitem[Houlsby et~al.(2019)Houlsby, Giurgiu, Jastrzebski, Morrone, De~Laroussilhe, Gesmundo, Attariyan, and Gelly]{houlsby2019parameter_adapter}
Neil Houlsby, Andrei Giurgiu, Stanislaw Jastrzebski, Bruna Morrone, Quentin De~Laroussilhe, Andrea Gesmundo, Mona Attariyan, and Sylvain Gelly.
\newblock Parameter-efficient transfer learning for nlp.
\newblock In \emph{International conference on machine learning}, pages 2790--2799. PMLR, 2019.

\bibitem[Hu et~al.(2022)Hu, Shen, Wallis, Allen-Zhu, Li, Wang, Wang, Chen, et~al.]{hu2022_lora}
Edward~J Hu, Yelong Shen, Phillip Wallis, Zeyuan Allen-Zhu, Yuanzhi Li, Shean Wang, Lu Wang, Weizhu Chen, et~al.
\newblock Lora: Low-rank adaptation of large language models.
\newblock \emph{ICLR}, 1\penalty0 (2):\penalty0 3, 2022.

\bibitem[Jia et~al.(2022)Jia, Tang, Chen, Cardie, Belongie, Hariharan, and Lim]{jia2022visual_vpt}
Menglin Jia, Luming Tang, Bor-Chun Chen, Claire Cardie, Serge Belongie, Bharath Hariharan, and Ser-Nam Lim.
\newblock Visual prompt tuning.
\newblock In \emph{European conference on computer vision}, pages 709--727. Springer, 2022.

\bibitem[Jiang et~al.(2023)Jiang, Mao, Huang, Lv, Zhao, and Zhou]{jiang2023rethinking_utuning}
Zeyinzi Jiang, Chaojie Mao, Ziyuan Huang, Yiliang Lv, Deli Zhao, and Jingren Zhou.
\newblock Rethinking efficient tuning methods from a unified perspective.
\newblock \emph{arXiv preprint arXiv:2303.00690}, 2023.

\bibitem[Jie and Deng(2023)]{jie2023facT}
Shibo Jie and Zhi-Hong Deng.
\newblock Fact: Factor-tuning for lightweight adaptation on vision transformer.
\newblock In \emph{Proceedings of the AAAI conference on artificial intelligence}, pages 1060--1068, 2023.

\bibitem[Kim et~al.(2024)Kim, Yang, Kim, Hong, and Park]{kim2024_hydra}
Sanghyeon Kim, Hyunmo Yang, Yunghyun Kim, Youngjoon Hong, and Eunbyung Park.
\newblock Hydra: Multi-head low-rank adaptation for parameter efficient fine-tuning.
\newblock \emph{Neural Networks}, 178:\penalty0 106414, 2024.

\bibitem[Krause et~al.(2013)Krause, Stark, Deng, and Fei-Fei]{krause20133d_standfordcar}
Jonathan Krause, Michael Stark, Jia Deng, and Li Fei-Fei.
\newblock 3d object representations for fine-grained categorization.
\newblock In \emph{Proceedings of the IEEE international conference on computer vision workshops}, pages 554--561, 2013.

\bibitem[Lan et~al.(2019)Lan, Chen, Goodman, Gimpel, Sharma, and Soricut]{lan2019albert}
Zhenzhong Lan, Mingda Chen, Sebastian Goodman, Kevin Gimpel, Piyush Sharma, and Radu Soricut.
\newblock Albert: A lite bert for self-supervised learning of language representations.
\newblock \emph{arXiv preprint arXiv:1909.11942}, 2019.

\bibitem[Li and Liang(2021)]{li2021_prefixtuning}
Xiang~Lisa Li and Percy Liang.
\newblock Prefix-tuning: Optimizing continuous prompts for generation.
\newblock \emph{arXiv preprint arXiv:2101.00190}, 2021.

\bibitem[Lian et~al.(2022)Lian, Zhou, Feng, and Wang]{lian2022scaling_ssf}
Dongze Lian, Daquan Zhou, Jiashi Feng, and Xinchao Wang.
\newblock Scaling \& shifting your features: A new baseline for efficient model tuning.
\newblock \emph{Advances in Neural Information Processing Systems}, 35:\penalty0 109--123, 2022.

\bibitem[Lin et~al.(2023)Lin, Wu, Yang, Huang, Huang, and Jin]{lin2023hierarchical_hst}
Weifeng Lin, Ziheng Wu, Wentao Yang, Mingxin Huang, Jun Huang, and Lianwen Jin.
\newblock Hierarchical side-tuning for vision transformers.
\newblock \emph{arXiv preprint arXiv:2310.05393}, 2023.

\bibitem[Liu et~al.(2021)Liu, Lin, Cao, Hu, Wei, Zhang, Lin, and Guo]{liu2021swin}
Ze Liu, Yutong Lin, Yue Cao, Han Hu, Yixuan Wei, Zheng Zhang, Stephen Lin, and Baining Guo.
\newblock Swin transformer: Hierarchical vision transformer using shifted windows.
\newblock In \emph{Proceedings of the IEEE/CVF international conference on computer vision}, pages 10012--10022, 2021.

\bibitem[Luo et~al.(2023)Luo, Huang, Zhou, Sun, Jiang, Wang, and Ji]{luo2023towards_RepAdapter}
Gen Luo, Minglang Huang, Yiyi Zhou, Xiaoshuai Sun, Guannan Jiang, Zhiyu Wang, and Rongrong Ji.
\newblock Towards efficient visual adaption via structural re-parameterization.
\newblock \emph{arXiv preprint arXiv:2302.08106}, 2023.

\bibitem[Maji et~al.(2013)Maji, Rahtu, Kannala, Blaschko, and Vedaldi]{maji2013fine_aircraf}
Subhransu Maji, Esa Rahtu, Juho Kannala, Matthew Blaschko, and Andrea Vedaldi.
\newblock Fine-grained visual classification of aircraft.
\newblock \emph{arXiv preprint arXiv:1306.5151}, 2013.

\bibitem[Nilsback and Zisserman(2006)]{nilsback2006visual_Oxford_Flowers102}
M-E Nilsback and Andrew Zisserman.
\newblock A visual vocabulary for flower classification.
\newblock In \emph{2006 IEEE computer society conference on computer vision and pattern recognition (CVPR'06)}, pages 1447--1454. IEEE, 2006.

\bibitem[Nilsback and Zisserman(2008)]{nilsback2008automated_flower_dataset}
Maria-Elena Nilsback and Andrew Zisserman.
\newblock Automated flower classification over a large number of classes.
\newblock In \emph{2008 Sixth Indian conference on computer vision, graphics \& image processing}, pages 722--729. IEEE, 2008.

\bibitem[Parkhi et~al.(2012)Parkhi, Vedaldi, Zisserman, and Jawahar]{parkhi2012cats_oxfoedpet}
Omkar~M Parkhi, Andrea Vedaldi, Andrew Zisserman, and CV Jawahar.
\newblock Cats and dogs.
\newblock In \emph{2012 IEEE conference on computer vision and pattern recognition}, pages 3498--3505. IEEE, 2012.

\bibitem[Recht et~al.(2019)Recht, Roelofs, Schmidt, and Shankar]{recht2019imagenet_v2}
Benjamin Recht, Rebecca Roelofs, Ludwig Schmidt, and Vaishaal Shankar.
\newblock Do imagenet classifiers generalize to imagenet?
\newblock In \emph{International conference on machine learning}, pages 5389--5400. PMLR, 2019.

\bibitem[Sung et~al.(2022)Sung, Cho, and Bansal]{sung2022_lst}
Yi-Lin Sung, Jaemin Cho, and Mohit Bansal.
\newblock Lst: Ladder side-tuning for parameter and memory efficient transfer learning.
\newblock \emph{Advances in Neural Information Processing Systems}, 35:\penalty0 12991--13005, 2022.

\bibitem[Tang et~al.(2024)Tang, Fu, Zhu, and Wu]{tang2024low_last}
Ningyuan Tang, Minghao Fu, Ke Zhu, and Jianxin Wu.
\newblock Low-rank attention side-tuning for parameter-efficient fine-tuning.
\newblock \emph{arXiv preprint arXiv:2402.04009}, 2024.

\bibitem[Touvron et~al.(2021)Touvron, Cord, Douze, Massa, Sablayrolles, and J{\'e}gou]{touvron2021training}
Hugo Touvron, Matthieu Cord, Matthijs Douze, Francisco Massa, Alexandre Sablayrolles, and Herv{\'e} J{\'e}gou.
\newblock Training data-efficient image transformers \& distillation through attention.
\newblock In \emph{International conference on machine learning}, pages 10347--10357. PMLR, 2021.

\bibitem[Van~Horn et~al.(2015)Van~Horn, Branson, Farrell, Haber, Barry, Ipeirotis, Perona, and Belongie]{van2015building_nabird_dataset}
Grant Van~Horn, Steve Branson, Ryan Farrell, Scott Haber, Jessie Barry, Panos Ipeirotis, Pietro Perona, and Serge Belongie.
\newblock Building a bird recognition app and large scale dataset with citizen scientists: The fine print in fine-grained dataset collection.
\newblock In \emph{Proceedings of the IEEE conference on computer vision and pattern recognition}, pages 595--604, 2015.

\bibitem[Vaswani et~al.(2017)Vaswani, Shazeer, Parmar, Uszkoreit, Jones, Gomez, Kaiser, and Polosukhin]{vaswani2017attention_transformer}
Ashish Vaswani, Noam Shazeer, Niki Parmar, Jakob Uszkoreit, Llion Jones, Aidan~N Gomez, {\L}ukasz Kaiser, and Illia Polosukhin.
\newblock Attention is all you need.
\newblock \emph{Advances in neural information processing systems}, 30, 2017.

\bibitem[Wah et~al.(2011)Wah, Branson, Welinder, Perona, and Belongie]{wah2011caltech_cub_dataset}
Catherine Wah, Steve Branson, Peter Welinder, Pietro Perona, and Serge Belongie.
\newblock The caltech-ucsd birds-200-2011 dataset.
\newblock 2011.

\bibitem[Wang et~al.(2018)Wang, Singh, Michael, Hill, Levy, and Bowman]{wang2018glue}
Alex Wang, Amanpreet Singh, Julian Michael, Felix Hill, Omer Levy, and Samuel~R Bowman.
\newblock Glue: A multi-task benchmark and analysis platform for natural language understanding.
\newblock \emph{arXiv preprint arXiv:1804.07461}, 2018.

\bibitem[Wang et~al.(2019)Wang, Ge, Lipton, and Xing]{wang2019learning_sketch}
Haohan Wang, Songwei Ge, Zachary Lipton, and Eric~P Xing.
\newblock Learning robust global representations by penalizing local predictive power.
\newblock \emph{Advances in neural information processing systems}, 32, 2019.

\bibitem[Wang et~al.(2023)Wang, Shi, Zhang, Li, Liu, Dai, Li, Xiong, and Tian]{wang2023adapting_snf}
Yaoming Wang, Bowen Shi, Xiaopeng Zhang, Jin Li, Yuchen Liu, Wenrui Dai, Chenglin Li, Hongkai Xiong, and Qi Tian.
\newblock Adapting shortcut with normalizing flow: An efficient tuning framework for visual recognition.
\newblock In \emph{2023 IEEE/CVF Conference on Computer Vision and Pattern Recognition (CVPR)}, pages 15965--15974. IEEE, 2023.

\bibitem[Xin et~al.(2024)Xin, Luo, Liu, Zhou, Cheng, Lee, Du, Wang, Chen, Liu, et~al.]{xin2024v_v_petf}
Yi Xin, Siqi Luo, Xuyang Liu, Haodi Zhou, Xinyu Cheng, Christina~E Lee, Junlong Du, Haozhe Wang, MingCai Chen, Ting Liu, et~al.
\newblock V-petl bench: A unified visual parameter-efficient transfer learning benchmark.
\newblock \emph{Advances in Neural Information Processing Systems}, 37:\penalty0 80522--80535, 2024.

\bibitem[Yang et~al.(2019)Yang, Dai, Yang, Carbonell, Salakhutdinov, and Le]{yang2019xlnet}
Zhilin Yang, Zihang Dai, Yiming Yang, Jaime Carbonell, Russ~R Salakhutdinov, and Quoc~V Le.
\newblock Xlnet: Generalized autoregressive pretraining for language understanding.
\newblock \emph{Advances in neural information processing systems}, 32, 2019.

\bibitem[Zaken et~al.(2021)Zaken, Ravfogel, and Goldberg]{zaken2021_bitfit}
Elad~Ben Zaken, Shauli Ravfogel, and Yoav Goldberg.
\newblock Bitfit: Simple parameter-efficient fine-tuning for transformer-based masked language-models.
\newblock \emph{arXiv preprint arXiv:2106.10199}, 2021.

\bibitem[Zhai et~al.(2019)Zhai, Puigcerver, Kolesnikov, Ruyssen, Riquelme, Lucic, Djolonga, Pinto, Neumann, Dosovitskiy, et~al.]{zhai2019large_vtab}
Xiaohua Zhai, Joan Puigcerver, Alexander Kolesnikov, Pierre Ruyssen, Carlos Riquelme, Mario Lucic, Josip Djolonga, Andre~Susano Pinto, Maxim Neumann, Alexey Dosovitskiy, et~al.
\newblock A large-scale study of representation learning with the visual task adaptation benchmark.
\newblock \emph{arXiv preprint arXiv:1910.04867}, 2019.

\bibitem[Zhang et~al.(2022)Zhang, Peng, Wu, Liu, Xiao, Fu, and Yuan]{zhang2022minivit}
Jinnian Zhang, Houwen Peng, Kan Wu, Mengchen Liu, Bin Xiao, Jianlong Fu, and Lu Yuan.
\newblock Minivit: Compressing vision transformers with weight multiplexing.
\newblock In \emph{Proceedings of the IEEE/CVF Conference on Computer Vision and Pattern Recognition}, pages 12145--12154, 2022.

\bibitem[Zhang et~al.(2024{\natexlab{a}})Zhang, Zhou, and Liu]{zhang2024neural_noah}
Yuanhan Zhang, Kaiyang Zhou, and Ziwei Liu.
\newblock Neural prompt search.
\newblock \emph{IEEE Transactions on Pattern Analysis and Machine Intelligence}, 2024{\natexlab{a}}.

\bibitem[Zhang et~al.(2024{\natexlab{b}})Zhang, Zhang, Gao, Zhang, Shutova, Zhou, and Zhang]{zhang2024gradient_gps}
Zhi Zhang, Qizhe Zhang, Zijun Gao, Renrui Zhang, Ekaterina Shutova, Shiji Zhou, and Shanghang Zhang.
\newblock Gradient-based parameter selection for efficient fine-tuning.
\newblock In \emph{Proceedings of the IEEE/CVF Conference on Computer Vision and Pattern Recognition}, pages 28566--28577, 2024{\natexlab{b}}.

\bibitem[Zhao et~al.(2024)Zhao, Wang, Zhao, Luo, Wang, and Shou]{zhao2024_sct}
Henry~Hengyuan Zhao, Pichao Wang, Yuyang Zhao, Hao Luo, Fan Wang, and Mike~Zheng Shou.
\newblock Sct: A simple baseline for parameter-efficient fine-tuning via salient channels.
\newblock \emph{International Journal of Computer Vision}, 132\penalty0 (3):\penalty0 731--749, 2024.

\bibitem[Zheng et~al.(2021)Zheng, Lu, Zhao, Zhu, Luo, Wang, Fu, Feng, Xiang, Torr, et~al.]{zheng2021rethinking}
Sixiao Zheng, Jiachen Lu, Hengshuang Zhao, Xiatian Zhu, Zekun Luo, Yabiao Wang, Yanwei Fu, Jianfeng Feng, Tao Xiang, Philip~HS Torr, et~al.
\newblock Rethinking semantic segmentation from a sequence-to-sequence perspective with transformers.
\newblock In \emph{Proceedings of the IEEE/CVF conference on computer vision and pattern recognition}, pages 6881--6890, 2021.

\bibitem[Zhou et~al.(2022)Zhou, Liu, Qiao, Xiang, and Loy]{zhou2022domain}
Kaiyang Zhou, Ziwei Liu, Yu Qiao, Tao Xiang, and Chen~Change Loy.
\newblock Domain generalization: A survey.
\newblock \emph{IEEE transactions on pattern analysis and machine intelligence}, 45\penalty0 (4):\penalty0 4396--4415, 2022.

\end{thebibliography}
}

\end{document}